\documentclass[10pt,times]{article}

\usepackage{graphicx}
\usepackage{natbib}
\setcitestyle{authoryear,open={(},close={)}}

\renewcommand{\cite}{\citet}

\usepackage[colorlinks,bookmarksopen,bookmarksnumbered,citecolor=red,urlcolor=red]{hyperref}

\usepackage{textcomp}

\begin{document}

\title{The Role of Edges in Line Drawing Perception}

\author{Aaron Hertzmann \\ Adobe Research\thanks{Preprint version; 
final paper to appear in \href{https://journals.sagepub.com/home/pec}{\textit{Perception}}.}}


\maketitle

\begin{abstract}
It has often been conjectured that the effectiveness of line drawings can be explained by the similarity of edge images to line drawings.  This paper presents several problems with explaining line drawing perception in terms of edges, and how the recently-proposed Realism Hypothesis of \cite{Hertzmann:2020:WDL} resolves these problems. There is nonetheless existing evidence that edges are often the best features for predicting where people draw lines; this paper describes how the Realism Hypothesis can explain this evidence.
\end{abstract}


\maketitle

\section{Introduction}

The human visual system seems finely optimized for the goal of helping us understand, navigate, and survive the natural world.  It has long been a mystery, then, why we are so good at understanding realistic line drawings, e.g., \cite{SayimCavanagh}. Why do we see drawings as depicting objects, and not just  as markings on a page or screen?  Even  observers who have never seen pictures before can recognize objects in line drawings, e.g., \cite{Jahoda1977,KennedyRoss,Kennedy1974,Deregowski:1989}. 

 One possible answer could be called the \textit{Lines-As-Edges Hypothesis}. In this hypothesis, drawings simulate natural images because lines are drawn where edges often occur in natural images.
The Lines-as-Edges Hypothesis appears to be explicitly proposed in only one academic paper, by \cite{SayimCavanagh}. Yet, anecdotally, it seems to reflect the ``conventional wisdom,'' at least amongst researchers of the author's acquaintance, many of whom seem to view edges as  sufficient explanation for line drawing.  Moreover, edge detection algorithms can often create very effective line drawings---in some studies, edge detection seems to perform best among existing methods for predicting how people create line drawings.

This paper points out several reasons why Lines-As-Edges alone cannot explain line drawing perception, and then attempts to explain the successes of edge-based line drawing algorithms.  
Specifically, it is argued that Lines-As-Edges is implausible because it asserts that the visual system ignores other image information in an arbitrary manner, and just for one very specific artistic style.   \cite{Hertzmann:2020:WDL} proposed an alternative explanation for line drawing perception, called the Realism Hypothesis. This paper explains how, in the Realism Hypothesis, edges may be good places to draw lines even if they do not explain perception. Finally, this paper makes  predictions of the behavior of cortical responses based on the Realism Hypothesis.

Key to this discussion is distinguishing \textit{where people draw lines} from \textit{why we perceive shape in lines}. It is tempting to conflate these concepts, e.g., to conclude that edge detection often produces good line drawings because we perceive lines as edges. This paper argues that the relationship between these concepts is subtler.

\section{Why Edges Alone Do Not Explain Line Drawing}

The Lines-as-Edges hypothesis arises from two compelling observations. First, since the pioneering experiments of \cite{Hubel1959,Hubel1968}, we know that the visual cortex includes cells that are responsive to edge patterns.  Second, if one runs an edge detection algorithm on a real image, then one often gets something that looks like a line drawing (Figure \ref{fig:davidLAE}B). 
\newcommand{\dogheight}{1.7in}
\newcommand{\davidlabel}[1]{\raisebox{\dogheight}{\textbf{\sf #1}}}
\newcommand{\davidgap}{-0.15in}
\begin{figure}
    \centering
    \davidlabel{A}\hspace{\davidgap}
    \includegraphics[height=\dogheight]{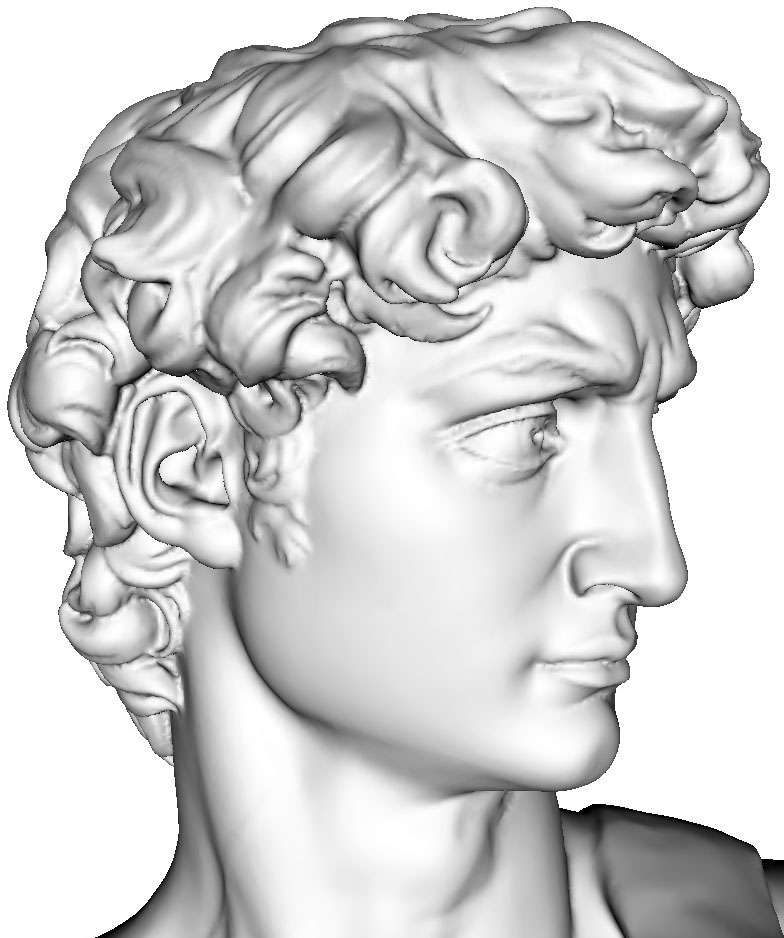}\hfill
    \davidlabel{B}\hspace{\davidgap}
    \includegraphics[height=\dogheight]{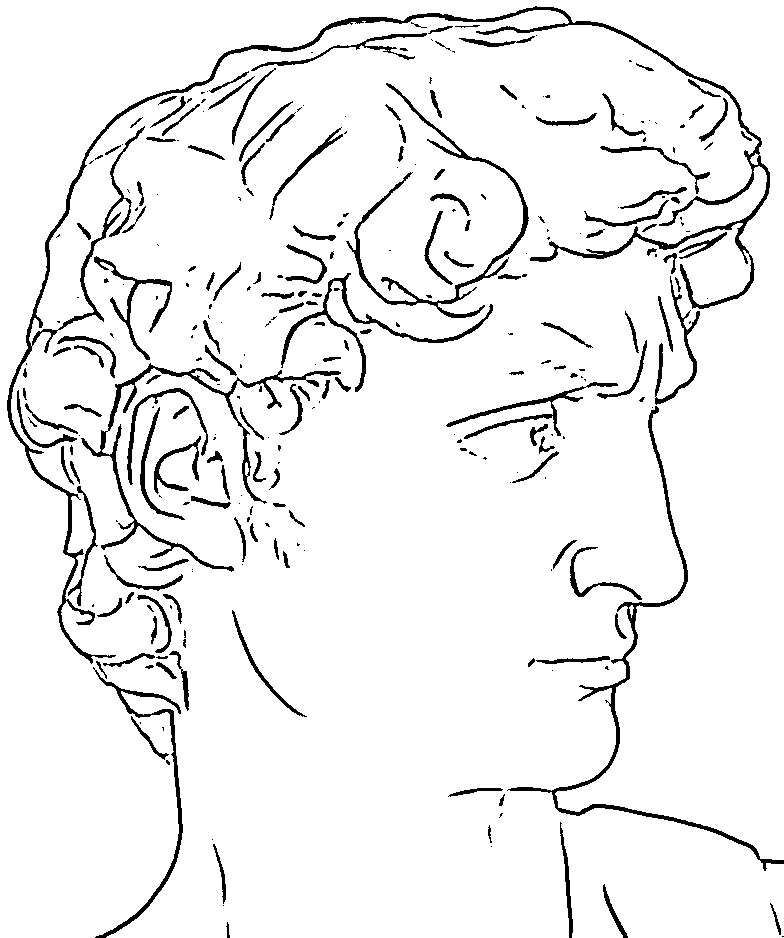}\hfill
    \davidlabel{C}\hspace{\davidgap}        \includegraphics[height=\dogheight]{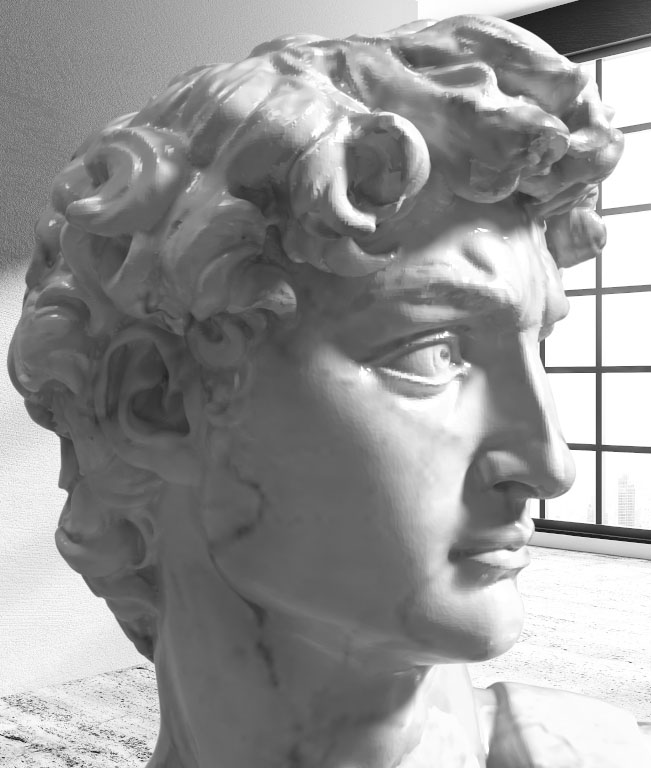}\hfill
        \davidlabel{D}\hspace{\davidgap}
\includegraphics[height=\dogheight]{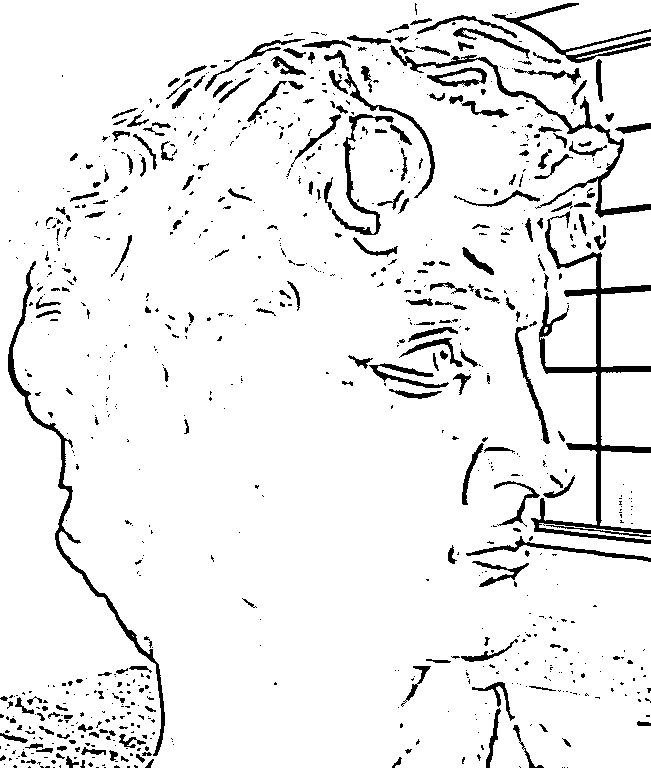}

    \caption{
Creating line drawings with edge detection.
        \textbf{Panel A:}
        A 3D model with white Lambertian material and headlight illumination.
        \textbf{Panel B:}
Edge detection on this image produces a plausible line drawing. 
 \textbf{Panel C:}
    Many realistic images contain contours that would not normally appear in a line drawing, due to effects like specular reflection, texture, and cast shadows. 
    \textbf{Panel D:} Edge detection on this image does not produce a plausible line drawing.
    (Edge images computed with Difference-of-Gaussians, see \cite{Marr:1980,Young};    
    \href{https://www.myminifactory.com/object/3d-print-2052}{Model} by Scan The World, \href{https://stock.adobe.com/images/interior-with-concrete-wall/303492332?asset_id=303492332}{background image} by peshkova)}
    \label{fig:davidLAE}
\end{figure}

One possible statement of the Lines-As-Edges hypothesis, based on \cite{SayimCavanagh}, is: lines are drawn at natural image edges, and thus activate the same edge receptor cells that the natural image would. The line drawing produces a cortical response like that of a natural image, and thus a viewer perceives the drawing and the photograph in roughly the same way.

There are four main problems with this hypothesis.

\paragraph{1. Edge images are not line drawings, and vice versa.}
The first problem is pointed out by
\cite{SayimCavanagh}: lines are often not drawn at image edges, and, conversely,
edge detection on natural images often produces images that do not look like line drawings. For example, cast shadows and texture can create contours that do not normally appear in line drawings (Figure \ref{fig:davidLAE}D).
\cite{SayimCavanagh} interpret this as evidence that the visual system somehow decides which edge responses to use and which ones to ignore, and they leave the nature of this process as an open question.
\newcommand{\mooneyheight}{1.2in}

\begin{figure}
    \centering
    \textsf{A} 
    \includegraphics[width=\mooneyheight]{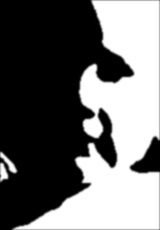}~~
    \textsf{B} 
    \includegraphics[width=\mooneyheight]{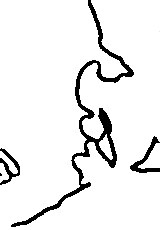} ~~~~~~
    \textsf{C} 
    \includegraphics[width=\mooneyheight]{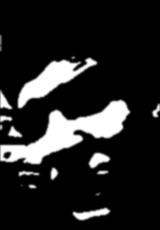} ~~
    \textsf{D} 
    \includegraphics[width=\mooneyheight]{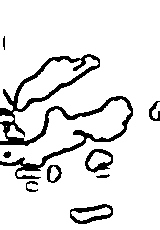}
\caption{Lines-As-Edges treats lines and edges as perceptually-equivalent. Hence, according to Lines-As-Edges, the image in Panel A should give the same shape percept as Panel B, as should C and D.  (Images A and C from \cite{Mooney2}, CC-BY.) }
    \label{fig:mooney}
\end{figure}

\paragraph{2. We can perceive more than edges.}
Lines-As-Edge supposes that the vision system ignores nearly all image information, other than a few isolated edge responses, in order to hallucinate 3D shape from a 2D surface containing a drawing. However, the human visual system is not solely an edge detector.  The primary visual cortex includes cells sensitive to line features and luminance, see \cite{Olshausen:2005}. We can tell the difference between a thin black line and the dark silhouette of an object against a light background, and these two things often look quite different to us (Figure \ref{fig:mooney}).

\paragraph{3. No higher-level rationale.}
Why would the visual system do this?  Some visual illusions can be explained by inductive biases that create unexpected outcomes from contrived stimuli, but aid perception in everyday situations. Lines-As-Edges does not assert any useful inductive bias, or any other explanation for why the visual system ought to be able to understand line drawings, that is, to perceive 3D shape from marks on a 2D page. 

\paragraph{4. It does not generalize.}
Representational visual art is more than just plain line drawing (Figure \ref{fig:drawings}A). We are capable of understanding shape and recognizing objects in a wide variety of visual styles, including hatching, stippling, Impressionist painting, Fauvism, cartoons, tile mosaics, needlepoint, gesture drawing, cave painting, and many more. 
None of these styles activate edge detectors in patterns matching those of natural images. Hence, these different styles would require separate explanations. 
This does not disprove Lines-As-Edges, which does not purport to explain any other styles. Nonetheless, it seems unsatisfying that it only applies to one very specific, basic line drawing style. Lines-As-Edges does not generalize even to slight variations in style, such as   sketchy, overdrawn, or highly-textured pencil strokes (Figure \ref{fig:drawings}B).  

\newcommand{\drawwidth}{1.4in}

\begin{figure}
\textsf{A} \\
       \includegraphics[width=\drawwidth]{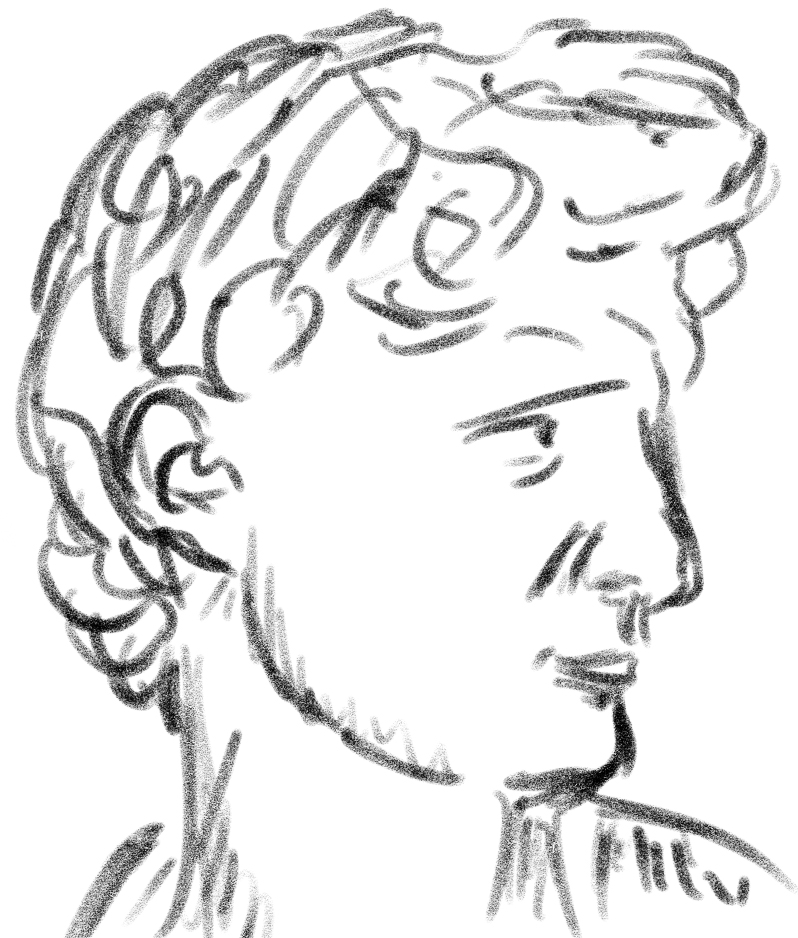}\hfill
       \includegraphics[width=\drawwidth]{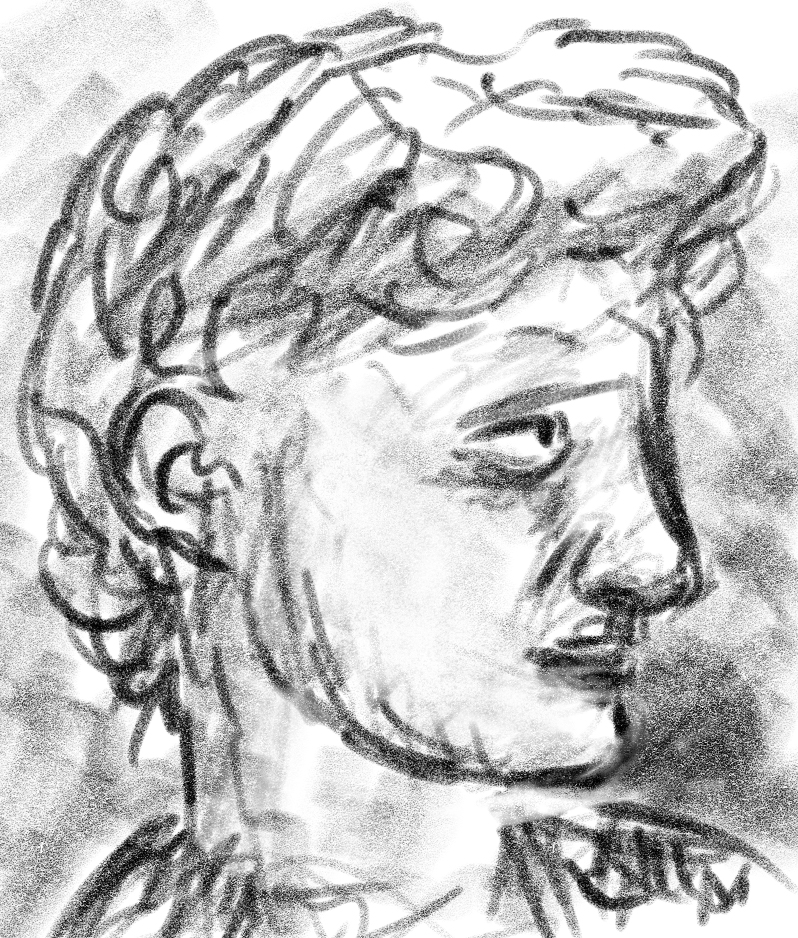}\hfill
          \includegraphics[width=\drawwidth]{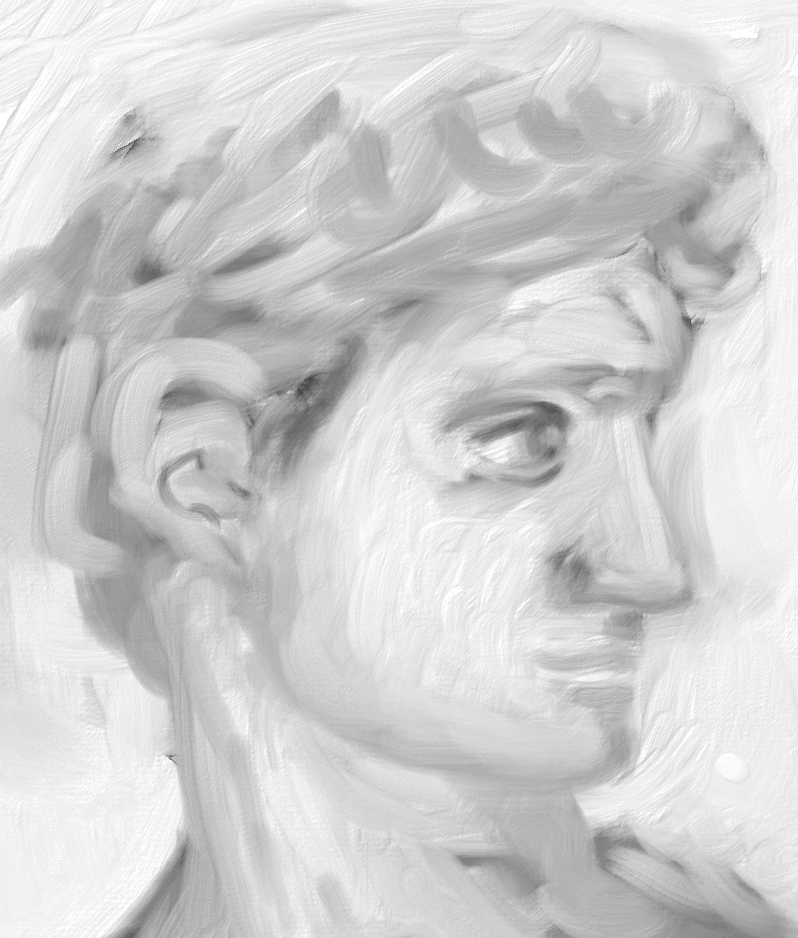}   \hfill
     \includegraphics[width=\drawwidth]{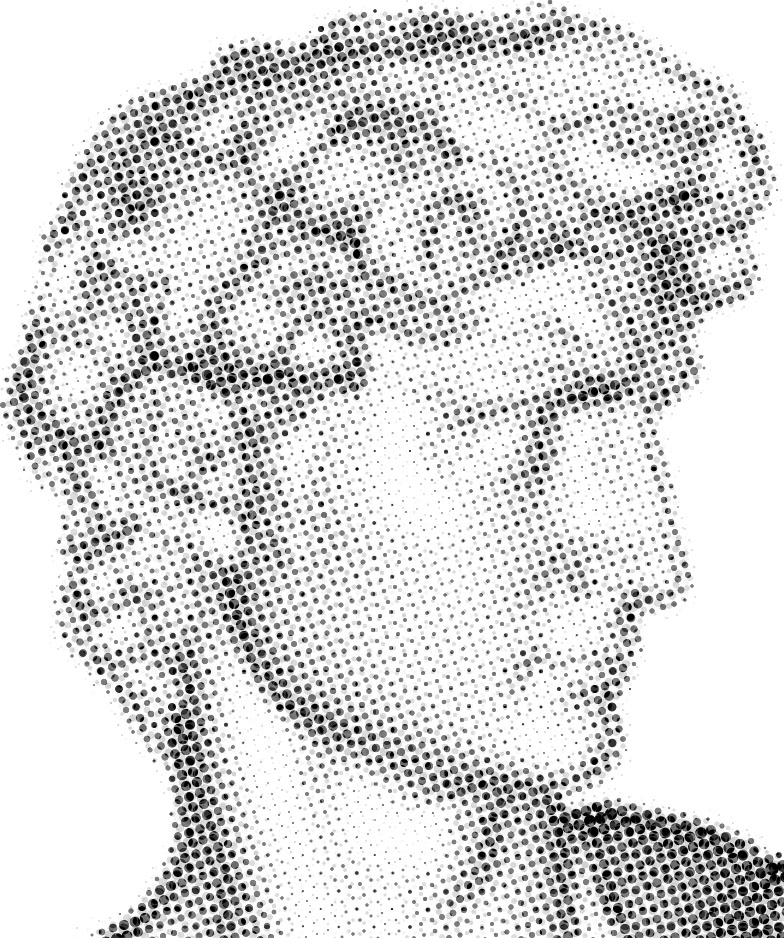}    \\[2ex]
     \textsf{B} \\
            \includegraphics[width=\drawwidth]{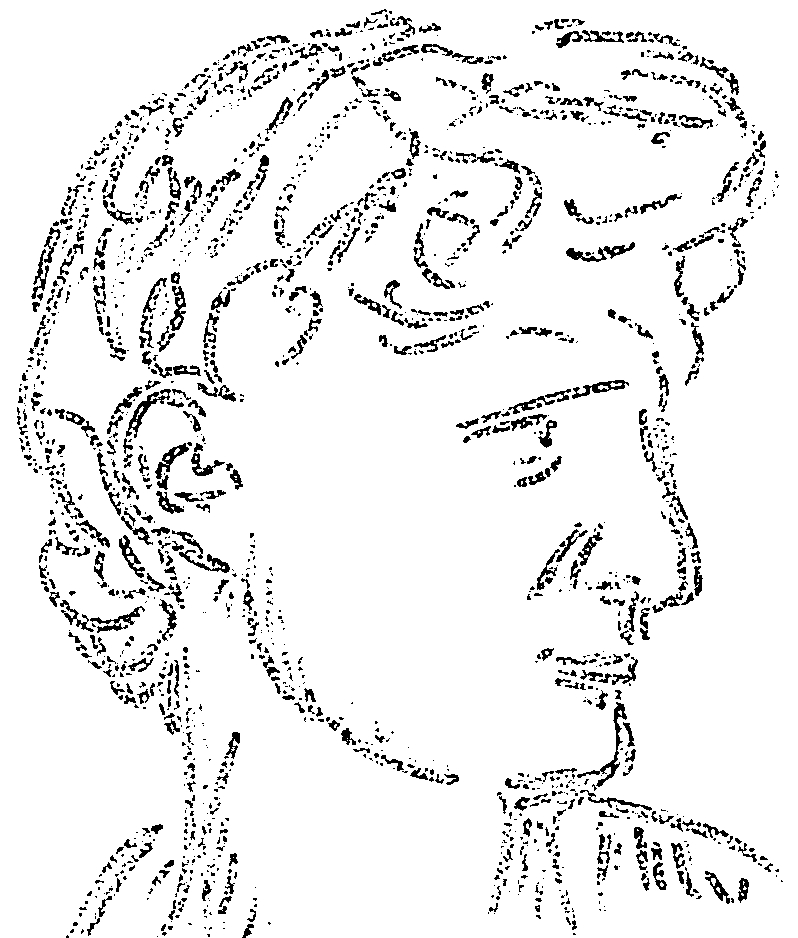}\hfill
            \includegraphics[width=\drawwidth]{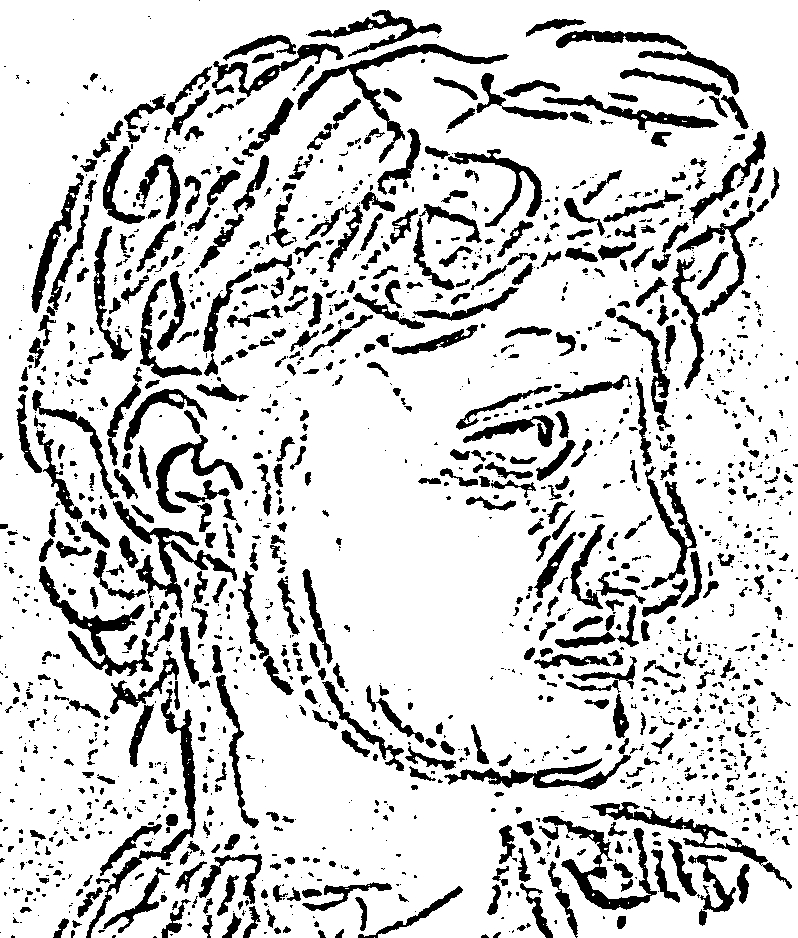}\hfill
          \includegraphics[width=\drawwidth]{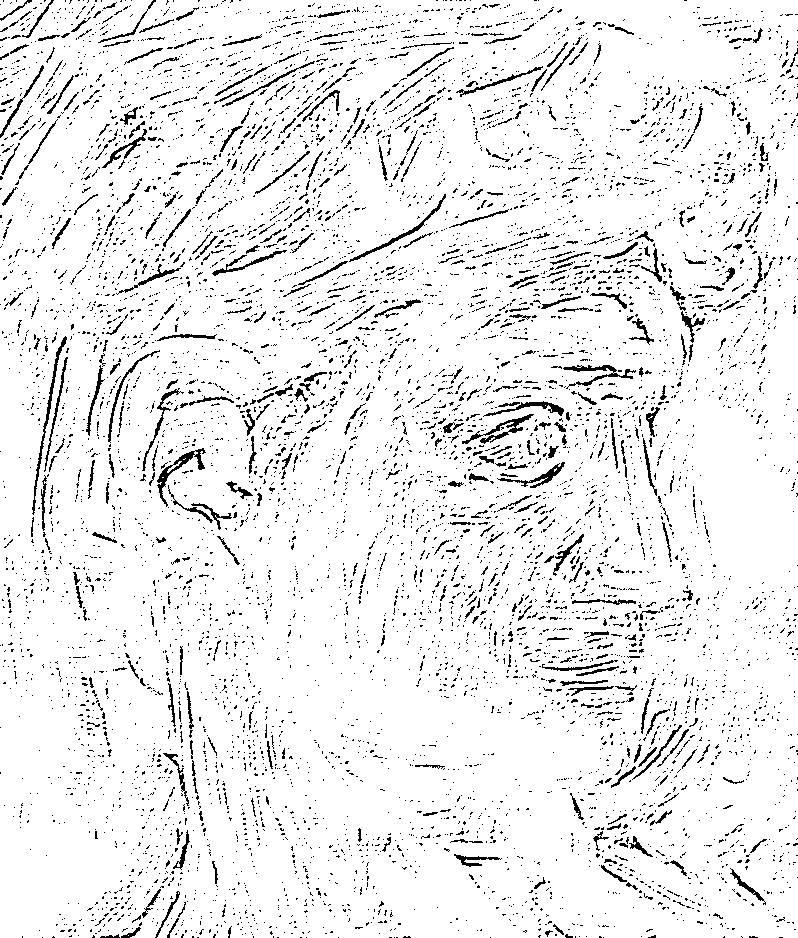}\hfill
     \includegraphics[width=\drawwidth]{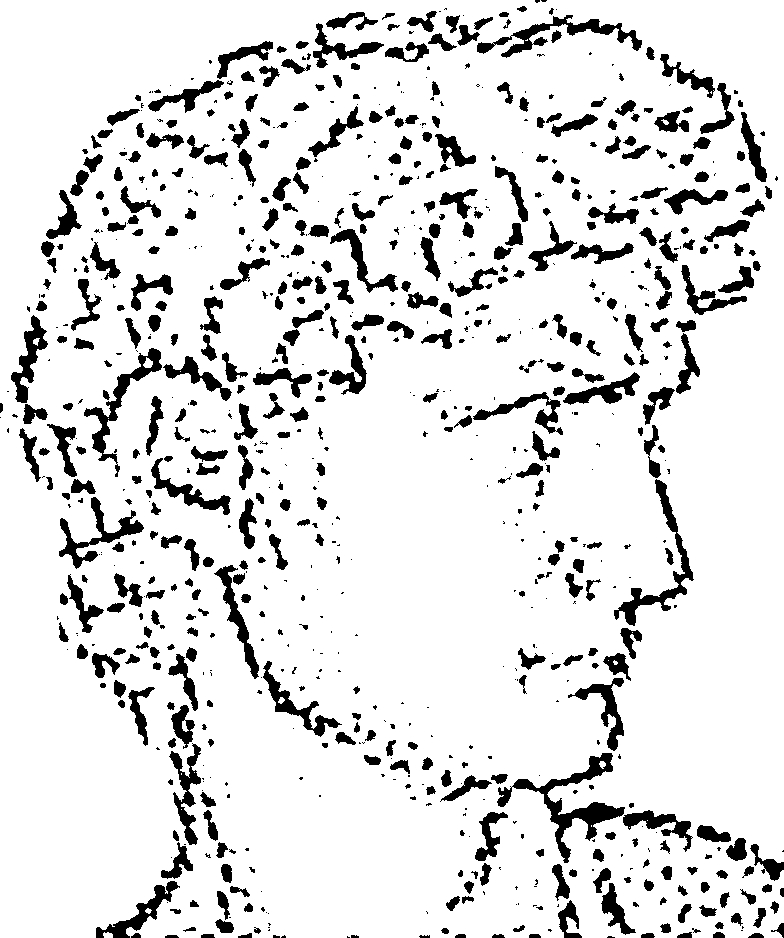}

    \caption{Different artistic styles can convey shape and appearance without preserving edges. Shading can convey tone that is not captured by edges. \textbf{Panel A:} Artistic renditions of the ``David'' model.  \textbf{Panel B:} Edges of the first row images. In each case, the parameters of the DoG filter were manually set to best capture the edges, with some cases using much larger blur factors than others, since it is unclear how to determine edge scale in Lines-As-Edges. Even when coarse-scale edges are preserved, the overall information contained in the image is different, especially with of shading. \label{fig:drawings}
}
\end{figure}

Suppose we had a perceptual theory that explained perception of many kinds of artistic styles. Then, most likely, this theory could explain line drawings as well, and Lines-As-Edges would not be needed.

\section{The Realism Hypothesis}

The Realism Hypothesis of \cite{Hertzmann:2020:WDL} can be summarized as follows: when viewing a line drawing in a basic style, the visual system perceives it as though it were a realistic image under a specific range of lighting and material conditions. 

\begin{figure}
    \centering
    \davidlabel{A}\hspace{\davidgap}
    \includegraphics[height=\dogheight]{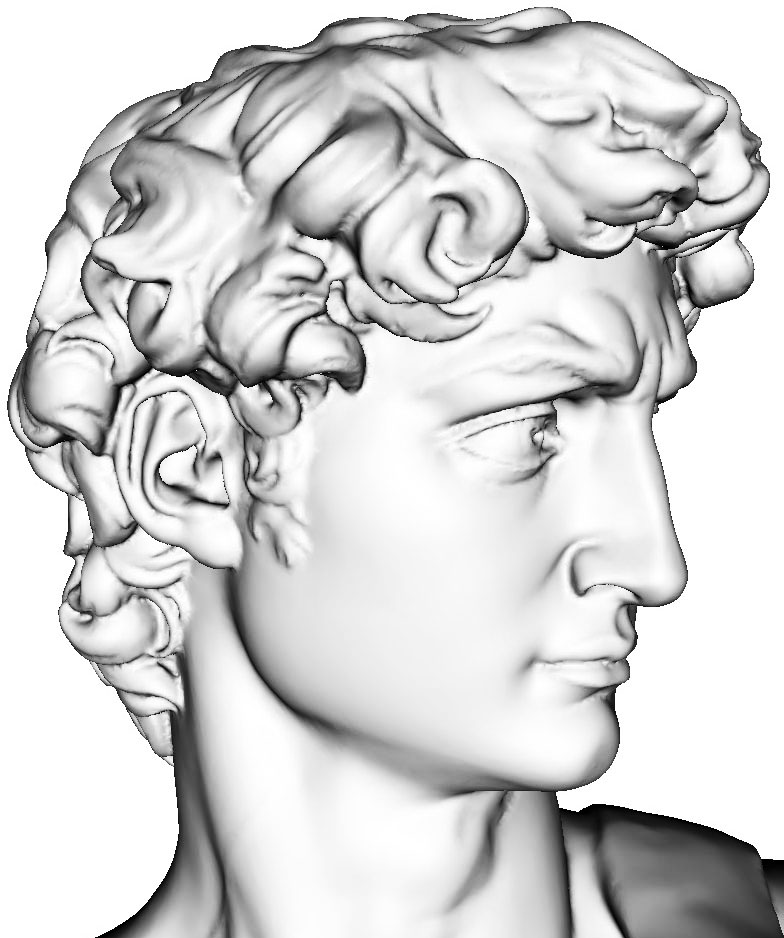}~~
    \davidlabel{B}\hspace{\davidgap}
    \includegraphics[height=\dogheight]{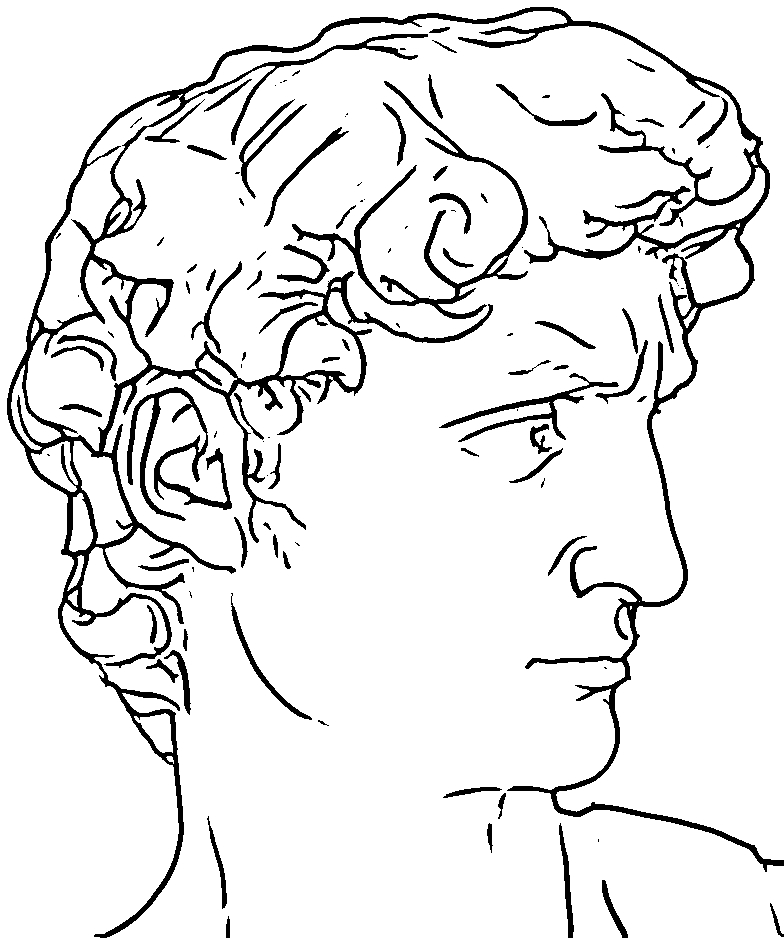}
    \caption{
Renderings in the Realism Hypothesis.
        \textbf{Panel A:}
        A 3D model with a white satin material, viewed with headlight illumination.
        \textbf{Panel B:}
        Valleys of the realistic rendering, detected using a valley-detection algorithm, see \cite{Lee:2007}.
    \label{fig:david_rh}}
\end{figure}

The following computer graphics rendering scenario demonstrates how realistic imaging can lead to line drawing-like images, see \cite{DeCarlo:2003,Lee:2007,Pearson:1985}.  Given a 3D scene, we cover everything in the scene with white materials, such as a white satin paint, or another ``rough'' material, e.g., \cite{OrenNayar}. 
We remove all light sources other than a headlight, that is, a point light source collocated with the viewer.  Viewing an object in this setup produces an image with characteristics similar to a line drawing; in particular, it has dark regions where lines would be drawn. An effective procedure for producing a drawing from this image is to draw black lines through the darkest regions in the rendering, as illustrated in Figure \ref{fig:david_rh}. The line drawing is visually similar to the shaded rendering, e.g., it has similar gray-levels and edges.  This setup hides cast shadows, textures, and other phenomena that would otherwise create spurious contours. Visual adaptation may also play a role, i.e., the visual system interprets flat white regions of a drawing as reflecting much more light than dark regions, e.g., \cite{vangorp}.

Unlike Lines-As-Edges, the Realism Hypothesis says nothing about the mechanics of human vision; it does not rely on knowledge of the visual cortex. It is, more generally, a statement of why any system that can understand real images---whether the system is biological or engineered by humans---ought to also be able to understand line drawings.

The Realism Hypothesis addresses the problems listed in the previous section more effectively than does Lines-As-Edges. 
Problems 1 and 2 do not apply, because the Realism Hypothesis does not rely on edges or other low-level features. It provides a high-level rationale (Problem 3), namely, that line drawing perception is a consequence of our ability to understand natural images. Finally, for Problem 4, the Realism Hypothesis could potentially generalize to other styles of art, since other styles of art can often be viewed as visual approximations to real images, e.g., \cite{HertzmannSBR}.

The Realism Hypothesis does not assume any sort of special processing for lines; lines are simply instances of attached shadows or highlights that happen to be curvilinear. This is consistent with evidence that line drawings are processed by the same neural pathways as natural scenes, see \cite{Walther}.

One could formulate a Modified Lines-As-Edges hypothesis based on the rendering setup above. This hypothesis would be: 
a line drawing activates the same edge receptor cells as would \textit{some} corresponding realistic rendering. For example, the visual system interprets the edges of Figure \ref{fig:davidLAE}B as if they were the edges of \ref{fig:davidLAE}A, not of \ref{fig:davidLAE}C.  
This resolves the problem pointed out by \cite{SayimCavanagh} and listed first in the previous section. 

This Modified Lines-As-Edges is compatible with the Realism Hypothesis. Both could be accurate. However, the Modified Lines-As-Edges does not seem to provide much advantage. Modified Lines-As-Edges says that \ref{fig:davidLAE}A and \ref{fig:davidLAE}B have similar edge features. The Realism Hypothesis also says that they are similar, without specifying in which features they are similar. They do have similar edges, but they also share similar overall luminance and gradients, and there is no particular reason to believe that only the edges matter.

\section{Line Drawing Algorithms and Their Evaluations}

We now turn to computer algorithms for creating line drawings from 3D models, and the studies that evaluate them, to see what they can tell us about perception of line drawing. 

Two leading principles have been proposed for line drawing algorithms.
The first is the principle of abstracted shading, see \cite{Pearson:1985,DeCarlo:2003,Lee:2007}, which the Realism Hypothesis is based on.
An important example is the Suggestive Contours algorithm, in which
a 3D model is rendered with  white Lambertian materials and a light source at the view position.  Abstracting a rendering of this scene, by drawing lines at dark valleys of the rendering (Figure \ref{fig:david_nose}B), produces Occluding Contours and Suggestive Contours. Abstracted shading with other lighting conditions leads to  other kinds of stylization, e.g., \cite{Lee:2007,DeCarlo:2007}. 

\newcommand{\davidwide}{1.8in}
\newcommand{\davidnosewide}{1.8in}
\newcommand{\xwide}{1.8in}
\newcommand{\spw}{0in}

\begin{figure}
    \centering
    \begin{tabular}{c@{ }c@{ }c@{ }c@{ }}
        {{\sf A.} Headlight} &  
        {\textsf{ B.} Occluding} &
    {\textsf{ C.} Apparent Ridges} \\[-0.5ex]
{Rendering} & {+ Suggestive Contours} & \\
    \includegraphics[width=\davidwide]{figures/david/David_diffuse.jpg} &
    \includegraphics[width=\davidwide]{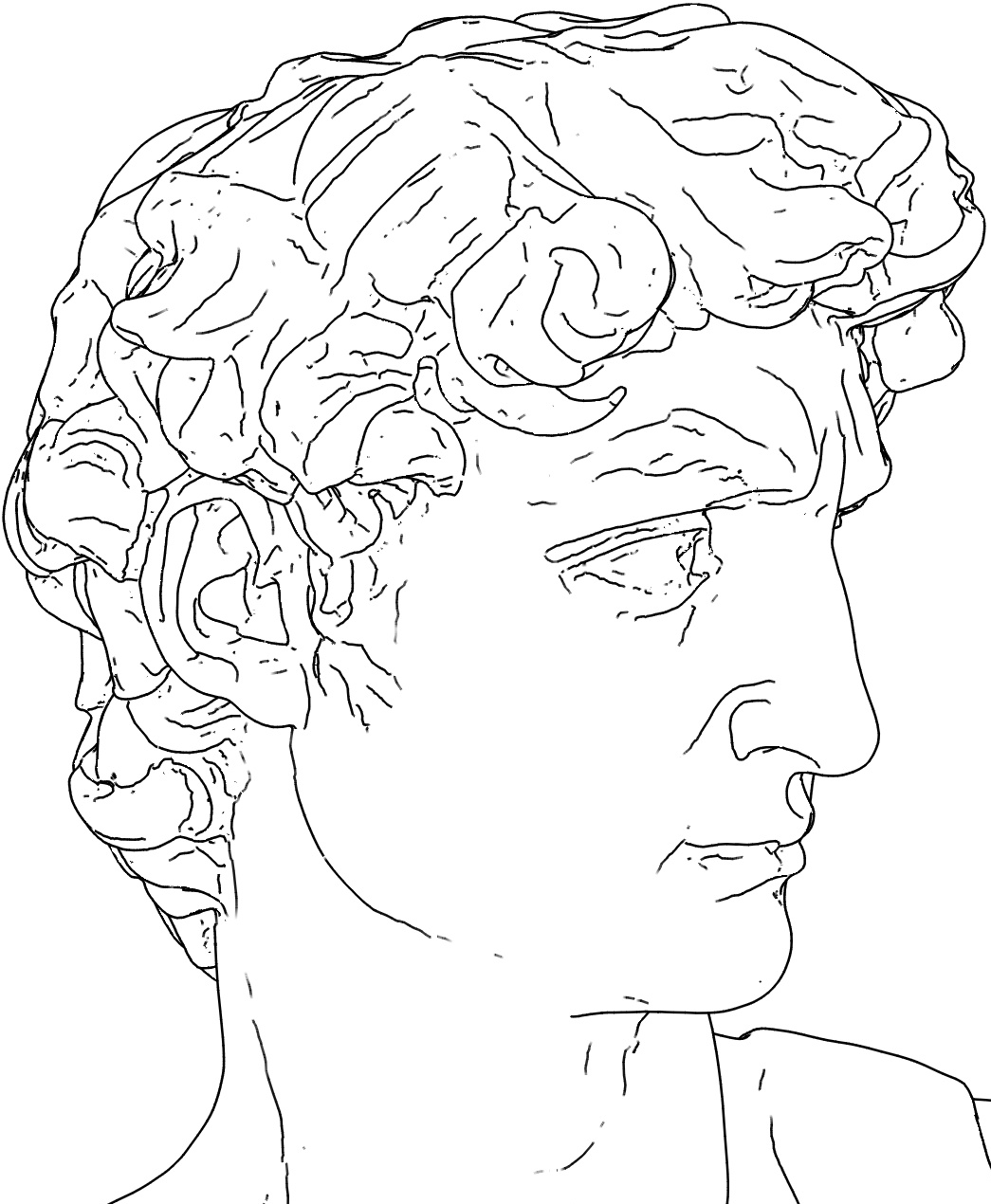} &
    \includegraphics[width=\davidwide]{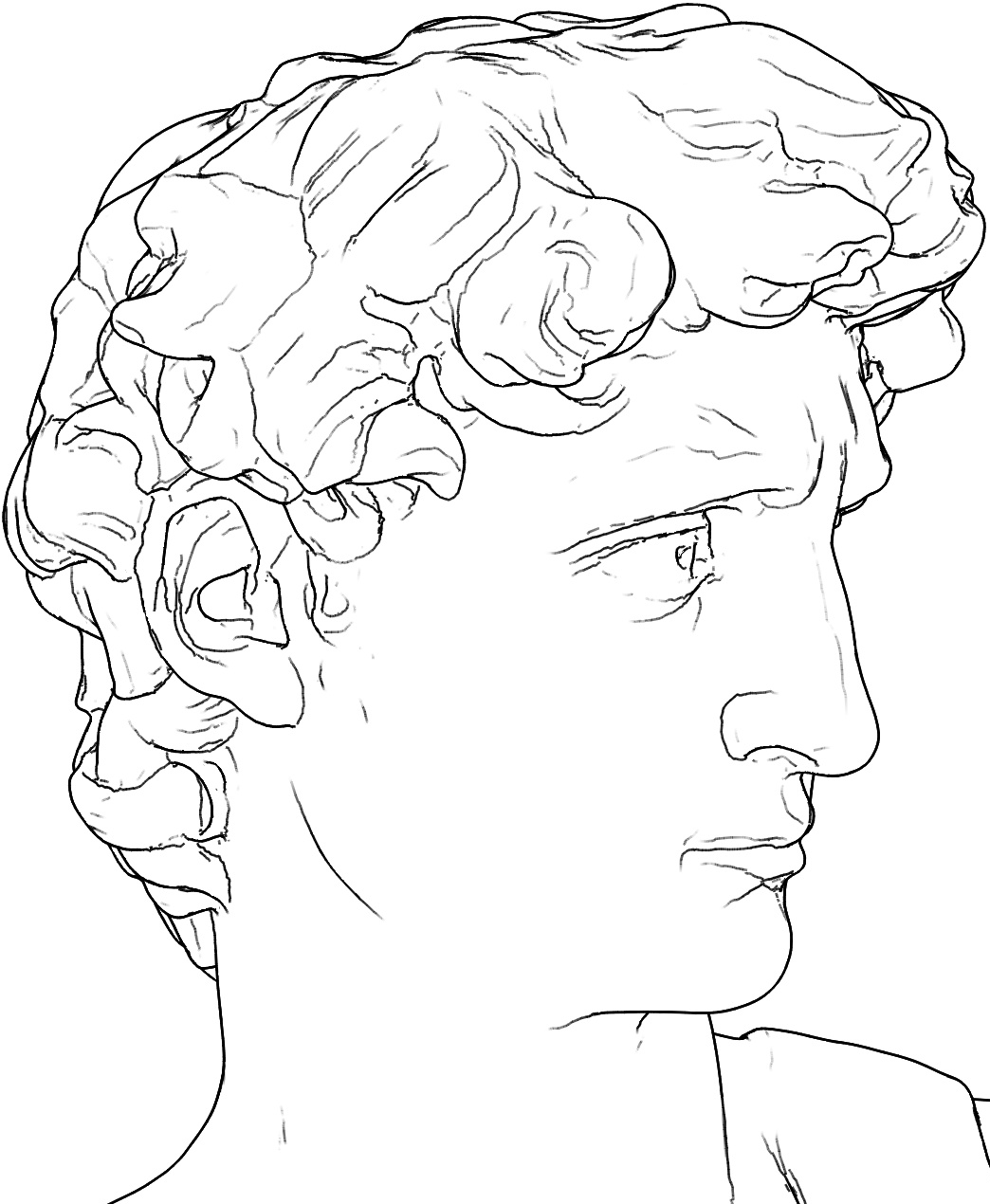} \\
        \includegraphics[width=\davidnosewide]{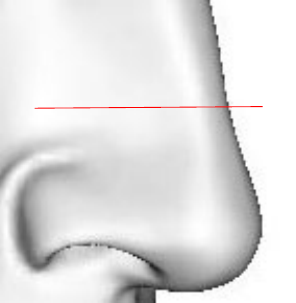} &
    \includegraphics[width=\davidnosewide]{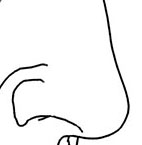} &
    \includegraphics[width=\davidnosewide]{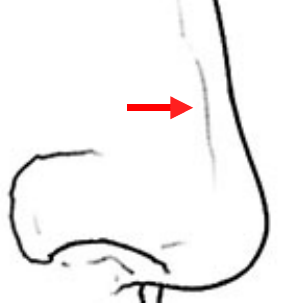} \\[2ex]
    \textsf{ D}~~
    \includegraphics[width=\xwide]{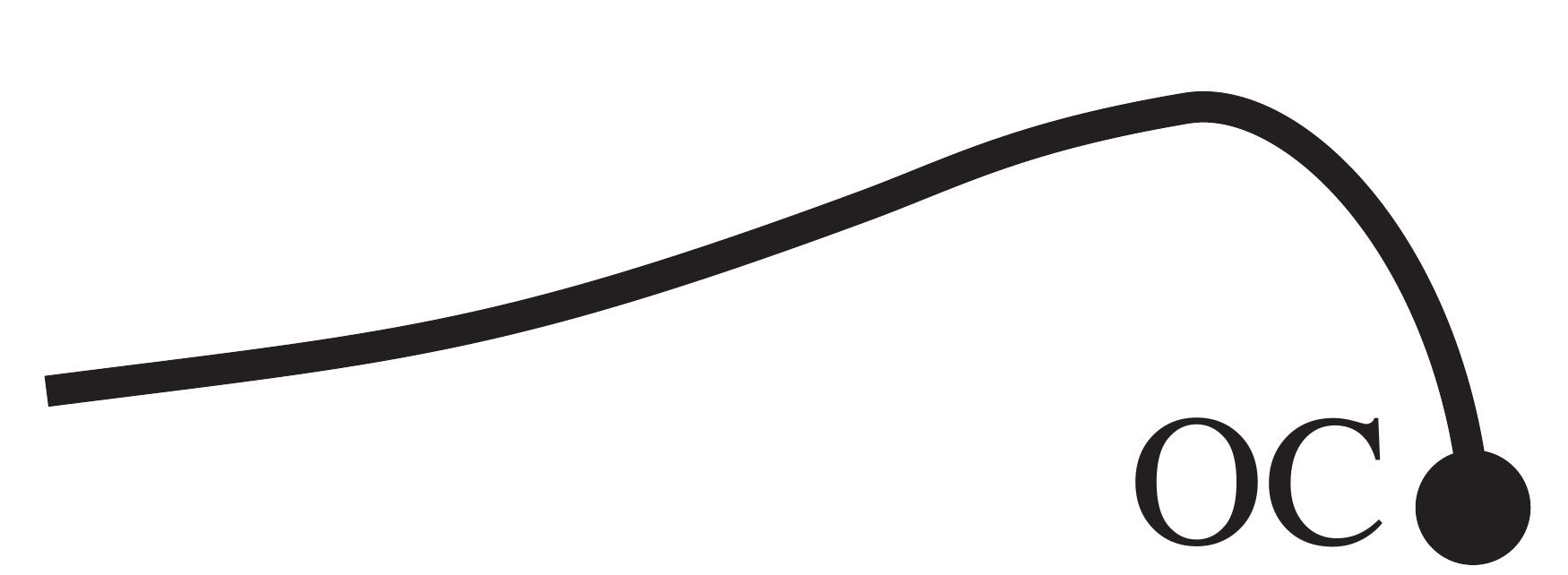}\hspace{\spw} &
    \textsf{ E}~~
    \includegraphics[width=\xwide]{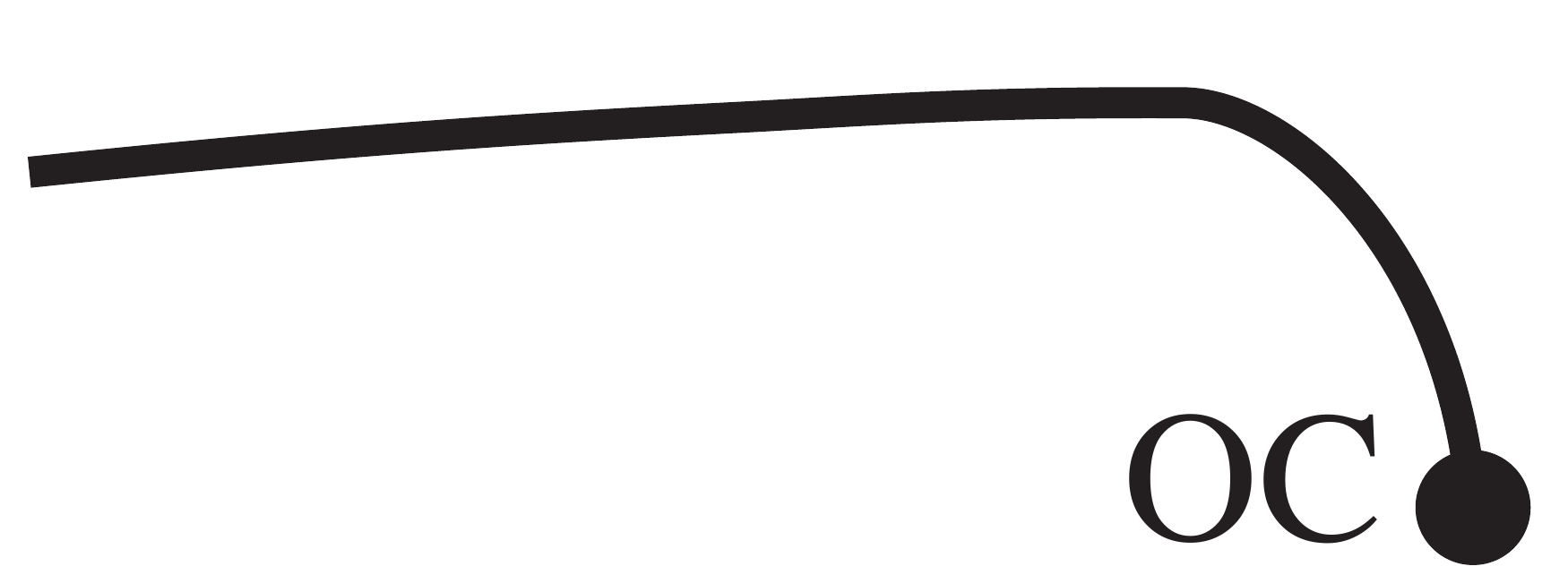}\hspace{\spw} &
    \textsf{ F}~~
    \includegraphics[width=\xwide]{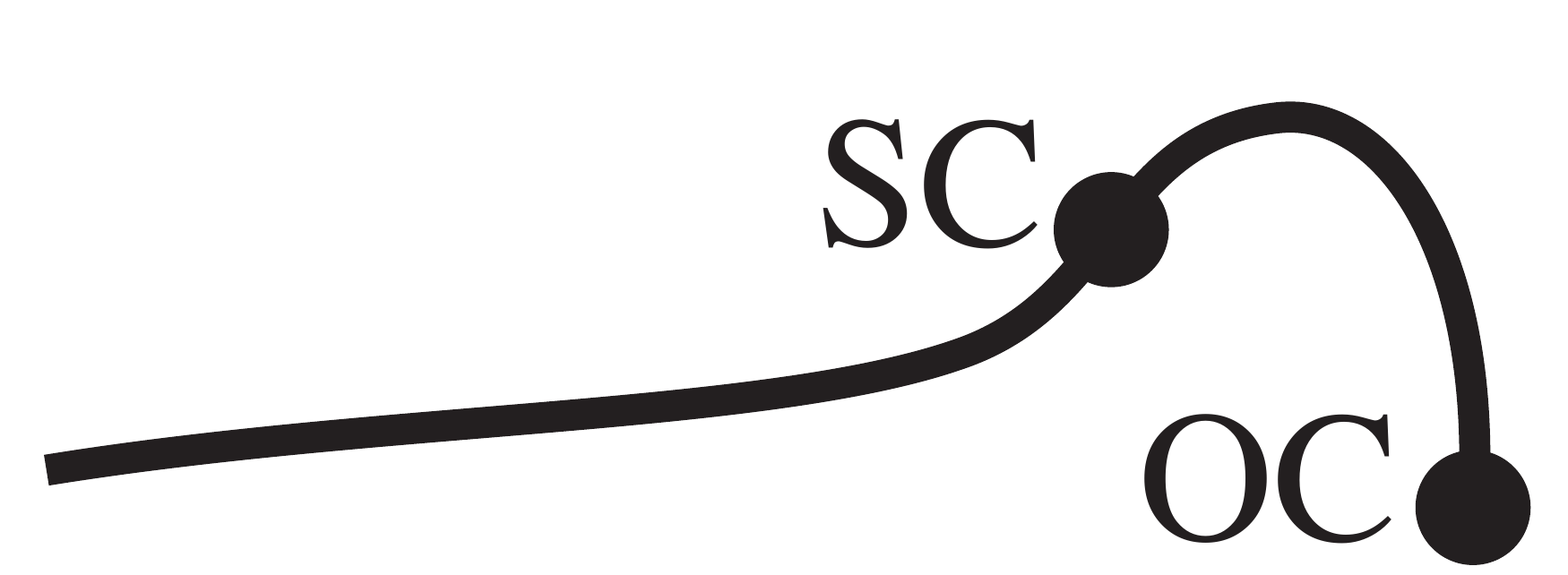}\hspace{\spw} \\
\end{tabular}
    \caption{Line drawing algorithms.
    Panels A, B, C: Three renderings of a 3D model.
    Panel D: possible cross-section of the 3D surface along the red line.
    Panels E, F: cross-sections of possible shape percepts of the drawings. There is an extra Apparent Ridge curve indicated by the red arrow. This curve on the nose may be perceived as a Suggestive Contour (SC) that implies a more pinched surface than the Suggestive Contour rendering does.
   (\href{https://www.myminifactory.com/object/3d-print-2052}{Model} by Scan The World, rendered with \href{https://github.com/fcole/qrtsc}{qrtsc} )
    }
    \label{fig:david_nose}
\end{figure}

The second principle is based on edge detection. An important example is the Apparent Ridges algorithm, see \cite{Judd:2007}.  Experimentally, Apparent Ridges occur at locations where large shading gradients stably occur under many different lighting conditions (Figure \ref{fig:david_nose}C).  Apparent Ridges subsume Occluding Contours.  Another variant predicts hand-drawn lines by applying edge detection to images of objects with white materials and some idealized lighting (Figure \ref{fig:davidLAE}A), as in \cite{Cole:2008}. 

Qualitatively, one may observe that the drawings in Figure \ref{fig:david_nose} agree in where most lines are placed; some differences are simply due to algorithmic nuances like thresholding parameters. One can find a few cases where the qualitative differences are more significant, e.g., see Figures 14 and 15 of \cite{DeCarlo:2012}, but these seem to be in the minority.

A few studies have  compared line drawing algorithms quantitatively in two different ways.  First, \cite{Cole:2008} asked human artists to create line drawings from 3D models, and
then evaluated which algorithms best predict where artists draw lines. They found that methods related to image edges and image gradients---including Apparent Ridges---were the best predictors of hand-drawn lines. However, there were shapes where Suggestive Contours performed better. 
\cite{Liu:2020:NC} observed similar outcomes with a broader range of comparisons.
Second, \cite{Cole:2009} used gauge studies to elicit viewers' perception of 3D shape from different line drawing algorithms.  They found that Apparent Ridges and Suggestive Contours convey shape with roughly-equivalent accuracy (see Table 1 in their paper). Apparent Ridges did slightly better on human-made models with many creases, and Suggestive Contours did a bit better on some other models. 

There are a few reasons for caution in interpreting these results. These studies generally compare curves in isolation, but one gets better scores by combining different kinds of curves. Importantly, the algorithms for integrating creases and Suggestive Contours have not been fully developed, whereas the Apparent Ridges implementation  does not handle Y-junctions where three creases meet.   The results may also depend heavily on the balance between organic and mechanical objects in the datasets, since the latter typically have many creases. Small differences in the evaluation results may also depend on various nuances in the evaluation metrics, such as sensitivity to precise positioning of curves. 

In the experiments described above, image edges and gradients seem to have a small but consistent advantage in predicting where people draw.  One might be tempted to treat this as evidence in favor of Lines-As-Edges.  However, both drawing principles exhibit good results when predicting where humans draw lines, and in producing drawings that convey 3D shape to viewers; qualitatively, each seems to have advantages or disadvantages in individual drawings.

\section{Drawing With Edges in the Realism Hypothesis}

How can we reconcile the ideas in the previous sections of this paper? That is, if edges do not explain line drawing perception, why are they so good at predicting line drawings?   Or, if abstracted shading explains perception, why doesn't abstracted shading explain where people draw lines?

This section outlines a way that the Realism Hypothesis, could explain why people often draw lines at gradients.  A key observation based on the previous section is: it does not seem that  \textit{where} people draw lines  tell us directly \textit{why} line perception works; the relationship may be more subtle.

Suppose that an artist's goal is to draw lines to accurately convey a 3D shape, for some notion of accuracy.
Basic line drawings omit shading information, making it impossible to convey shape in precise detail.  
Moreover, suppose the artist knows that the viewer will interpret lines as if they were Suggestive Contours. The obvious choice would be to draw the Suggestive Contours.  But this might not lead to the most accurate shape percept.

Consider the line indicated by the arrow in Figure \ref{fig:david_nose}C; similar lines appear in the artist-drawn examples in Figure 3 of \cite{Judd:2007}.   
This line can be interpreted as exaggerating the curvature of the nose, in order to show that the nose is not flat. This line does not appear in the Suggestive Contour rendering, and, without the line, this region is empty and flat (Figure \ref{fig:david_nose}E).  Adding the line implies a ``pinched'' surface, with higher curvature than the true surface (Figure \ref{fig:david_nose}F). However, this  exaggerated shape is arguably a more accurate percept than flat version.

The above example could be generalized to the following rule: in areas of high curvature, add a shadow line to accentuate the perceived curvature, even if no Suggestive Contour appears there.  Since shading gradients are likely to occur perpendicular to directions of high curvature, this is a rule that draws lines at likely image edges, i.e., at Apparent Ridges. Another example of this is in Figure 7 of \cite{Cole:2009} (right column, boxed region).

The examples in this section show that, if abstracted shading explains perception of line drawing, drawings could nonetheless correlate more strongly with image edges than with shading.

\section{Possible Predictions and Experiments}

The discussion here shows that the relationship between the visual cortex, perception of drawings, and line drawings is subtler than it seems at first. The Realism Hypothesis addresses these nuances, but as such it involves much conjecture and  possible variants.  There are several types of experiments that could be performed to refine the theory, to fill in the details, and to test the hypothesis.

One kind of evidence comes from studies on observers who have never seen line drawings, e.g., \cite{Kennedy1974,KennedyRoss,Jahoda1977,Deregowski:1989}.   For example, one might determine whether untrained observers can understand accurate line drawings based on creases, or perform figure-gauge studies. However, such studies may be infeasible; the above authors describe many difficulties in performing studies with untrained observers.

The Realism Hypothesis can also make predictions about the human visual cortex responses. 
Specifically, it predicts that the cortical responses to a line drawing should be very similar to those of a corresponding natural image, e.g., Figure \ref{fig:david_rh}A,B). This is borne out by the experiments of \cite{Walther} on categorical responses. 
\cite{MorganMRI} find that  line drawing features are highly correlated with cortical responses as compared to natural photographs: the Realism Hypothesis predicts that the images in Figure \ref{fig:david_rh}A,B would have similar correlations to cortical responses as measured in that experiment.

Computer vision models provide a third route for experimentation.
One prediction made in \cite{Hertzmann:2020:WDL} is that a computer vision algorithm that can understand shape in natural images ought to be able to understand shape in line drawings as well, and a successful test is shown in that paper.
Recent artificial neural networks accurately model cortical responses in human vision, e.g., \cite{yamins}, at least in aggregate. A second prediction is that an accurate neural model should have similar cortical responses to a line drawing as to they do to a corresponding realistic image.
Moreover, probing how these networks operate, e.g., as in \cite{ZeilerFergus,bau2020units}, could provide more precise understanding of how precisely line drawings are interpreted.

\bibliography{contour_tutorial,perception}

\end{document}